# DO WE NEED HIGHER-ORDER PROBABILITIES AND, IF SO, WHAT DO THEY MEAN? *

**Judea Pearl**
Cognitive Systems Laboratory
Computer Science Department
UCLA, Los Angeles, CA. 90024-1600

## ABSTRACT

The apparent failure of individual probabilistic expressions to distinguish uncertainty about truths from uncertainty about probabilistic assessments have prompted researchers to seek formalisms where the two types of uncertainties are given notational distinction. This paper demonstrates that the desired distinction is already a built-in feature of classical probabilistic models, thus, specialized notations are unnecessary.

## Introduction

There is no doubt that people make a distinction between sure and unsure probabilistic judgments. For example, everyone would agree that a typical coin has a 50% chance of turning up head, while most people would hesitate to assign a definite probability to a coin produced in a gambler's basement. For that reason we sometimes feel more comfortable assigning a range, rather than a point estimate of uncertainty, thus expressing our ignorance, doubt or lack of confidence in the judgment required. We may say, for example, that the probability of the coin turning up head lies somewhere between 60% and 40%, having no idea whether or how the coin was biased.

The apparent failure of individual probabilistic expressions to distinguish between uncertainty and ignorance, certainty and confidence, have swayed many researchers to seek alternative formalisms, where confidence measures are provided explicit notation (Shafer, 1976). We first describe the difficulties associated with representing confidence measures in classical probability theory, using the so called "higher-order probabilities" or "probabilities of probabilities". We then demonstrate, using an interpretation advanced by de Finetti (1977), how the causal networks formulation of probabilities facilitates the representation of confidence measures as an intrinsic part of one's knowledge system, requiring no specialized notation nor the use of higher-order probabilities.

## The Semantics of Probabilities of Probabilities

Traditional probability theory insists that probabilities be assigned strictly to propositions about factual events, namely, to sentences whose truth value can, at least in principle, be verified unequivocally by empirical tests. Following this tradition, we find it hard to assign probabilities to probability distributions themselves. The truth of probabilistic statements about any specific event cannot be ascertained nor falsified empirically; once we observe the uncertain

---

\* This work was supported in part by National Science Foundation Grant DCR 83-13875

event, i.e., the object of the statement, its probability becomes either one or zero, trivially ruling out all intermediate values. Even if we adopt the frequency interpretation of probability, the difficulty remains unresolved. If by first-order probabilities we mean the proportion of times that a given event occurs, to define second-order probabilities we must count the proportion of times that a given proportion occurs -- a hopeless task by any standard.

Conceptual difficulties also plague the subjective interpretation of second-order probabilities. If the statement $P(A) = p$ stands for one's mental state of certainty regarding the truth of $A$, then there is only one such state at any given time, and, again, what do we mean by a state of certainty in a state of certainty? If our mental machinery is equipped with some hypothetical "thermometer" with which we measure the certainties of various statements, then, presumably, the thermometer is designed to take as inputs truths of propositions. How can we then apply it to suddenly take as inputs degrees of certainty, namely, outputs of some other "thermometers"? Or, perhaps we have two types of thermometers, one for measuring certainty of propositions, the other for measuring reliability of thermometers?

In certain cases we find it convenient to envision Nature as a two-level lottery process: (1) a probability distribution $P$ is chosen at random from some urn containing a hidden mixture of such distributions, (2) the chosen distribution is permitted to uncover its identity via empirical observations. Confidence levels, in this picture, are means for expressing uncertainty relative to the first process, i.e., the selection of the urn. However, it is well known that there is no empirical procedure by which one can distinguish the two-level-lottery model from some equivalent single-level model and, conversely, the betting behavior of a rational agent (relative a single event) that believes in a two-level model cannot be distinguished from that of another rational agent believing in an equivalent single-level model. If I am forced to make a bet on the next outcome of the suspect coin then, having no idea which way the coin might possibly be biased, I would have no choice but take the expected value of the distribution and bet as though the coin was fair.

It is important to note that the nature of these difficulties is not mathematical but epistemological, reflecting the absence of clear empirical basis for assigning confidences to probabilities. Mathematically speaking, it is possible to construct axiomatic theories of higher-order probabilities that will exhibit some desired properties (Domotor, 1981; Gaifman, 1986). This has a respectable tradition (Fisher, 1957; Good, 1965) and has culminated in the nonprobabilistic but mathematically elegant approach of Dempster (1965) and Shafer (1976).

The utility of such theories lies in how well they account for the origin of partial confidence and how well they predict systematic fluctuations in perceived confidence levels as new information arrives. Therefore, a theory of confidence measures may remain purely tautological until a clear empirical semantic is attributed to its primitive parameters (e.g., the basic ''mass'' assignments in Dempster-Shafer theory), i.e., until the theory embraces clear procedures for determining how different streams of raw observations give rise to different confidence intervals[1]. Thus, the questions remain: what do people mean when they assign confidence intervals to probabilistic sentences? What empirical and/or procedural information is conveyed by such intervals? how these intervals expand and contract in light of new information? and, should we carry these intervals in AI systems or simply dismiss them as another peculiarity of unfortunate homosapiens.

We shall now cast the semantics of second-order probability statements within the framework of classical, first-order probabilistic theory.

Our starting point is the claim that probabilistic statements such as $P(A) = p$ are in themselves empirical events, of no lesser stature than other sentences reporting empirical observations. While not referring to an event open to full public scrutiny, these statements do, nevertheless, report outcomes of genuine experiments, namely, the mental procedures invoked in assessing the belief of a given proposition $A$. Thus, stating "event $A$ has a chance $p$ of occurring" is equivalent to stating "the mental event of computing the likelihood of $A$ has produced the outcome $p$".

Having endowed probabilistic statements with event status neutralizes the syntactic objection against writing sentences such as $P[P(A) = p]$. Both the square and round brackets enclose arguments of the same type, namely, empirical events. True, the latter event is external while the former personal. However, in the post-behaviorism era of computational cognition this distinction no longer represents a barrier to useful semantics; having adopted a computational model of knowledge representation (e.g., semantic networks, causal models) should permit us to specify the mental procedures involved in belief assessments with the same clarity and precision that we specify experimental procedures in a laboratory setting. What remains to be done is first, to explain what renders the event $P(A) = p$ an unknown, random event, rather than a fixed outcome of a stable procedure. Second, to explicate more precisely the mental procedures involved in making the two assessments, $P(A)$ and $P[P(A) = p]$, identify their empirical content and specify their pragmatic computational role.

## De Finetti's Paradigm of Uncertain Contingencies

A paradigm answering the first question has been suggested by de Finetti (1977) and has been guiding the Bayesian interpretation of confidence measures for over a decade (Spiegelhalter, 1986, Heckerman and Jimison, 1987). The basic idea is that the event $P(A) = p$ is perceived to be a random variable whenever the assessment of $P(A)$ depends substantially on the occurrence or non-occurrence of some other events in the system. In the words of de Finetti:

---

(1) Dempster-Shafer theory for example, predicts that all confidence intervals should disappear whenever one possesses a precise probabilistic model of a domain and whenever the available evidence consists of confirmations and denials of events predicted by the model. Thus, the theory fails to capture the obvious lack of confidence in the judgment "the coin will turn up head with probability 50%" if the coin was produced by a defective machine -- precisely 49% of its output consists of double-head coins, 49% are double-tail coins, and the rest are fair. Neither will the theory explain why we normally regain confidence in the judgment above as soon as we are told that, in the past two trials, the coin's outcomes were a tail and a head. Apparently, the belief intervals encoded in Dempster-Shafer's theory have a totally different origin than those portrayed by the example above.

> "The information apt to modify the probability assessed for an event $E$ - in so far as the observation of $H_i$ makes us change from $P(E)$ to $P(E \mid H_i)$ - can make us view the $H_i$'s as sort of "noisy" signals concerning the occurrence and non-occurrence of the event $E$."

Adopting this interpretation, we shall further show that the procedure involved in the assessment of $P[P(A) = p]$ is no different than that involved in the assessment of $P(A)$ and, moreover, that the very information used for calculating $P(A)$ is sufficient for calculating the confidence interval associated with the statement $P(A) = p$. Thus, the notions of ignorance and doubt are intrinsic and indigenous to classical probabilistic formulation; no second-order probabilities nor specialized notational machinery are required to reinstate them where they flourish so naturally.

We shall first illustrate the basic idea using the two-coin example, coin-1 being a typical off-the-street coin, and coin-2 having been minted in the basement of a notoriously unscrupulous gambler. Let $E_i$, $i = 1, 2$, stand for the statement "coin-$i$ is about to turn head". To say that we are unsure about the probability of $E_2$ means that our belief in it is extremely susceptible to change in light of further evidence. More specifically, being unsure about the probability $P(E_2)$ means that one is aware of a set of CONTINGENCIES, each of which is rather likely to be true and , if true, would substantially alter our assessment of the probability in the outcome of the next toss. For example, since coin-2 has been given to us by a disreputable gambler, we may entertain the following contingencies:

$C_1$ - The coin is fair ($p = .5$).

$C_2$ - The coin was loaded toward head ($p = .6$)

$C_3$ - The coin was loaded toward tail ($p = .4$)

If we attribute a 20% chance to the possibility that the coin has been tampered with, then our assessment of the likelihood of these three conditions can be summarized by a probability vector $P(C_1) = .8$ $P(C_2) = .10$, $P(C_3) = .10$ (see Figure 1(a)), and the distribution of $P(E_2)$ will assume the form given in figure 1(b). In some applications it might also be useful to characterize this distribution with several of its attributes, for example, the range [0.40, 0.60], the mean 0.50, or the variance $\sigma^2 = 0.002$.

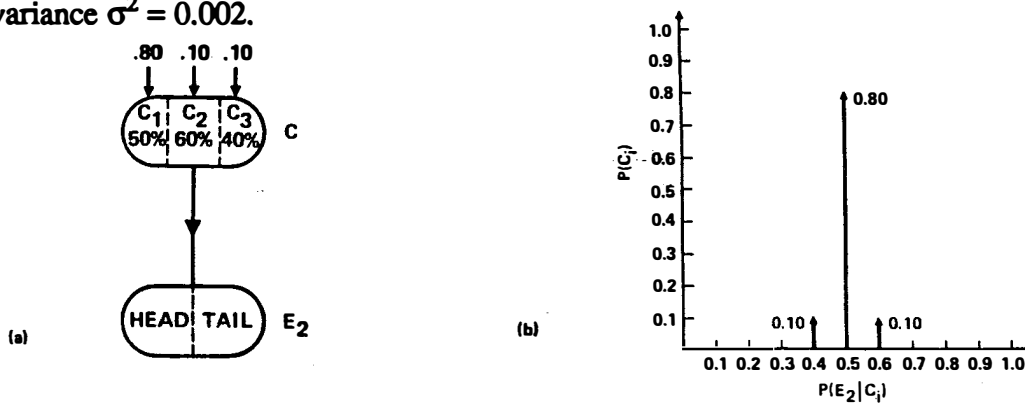

**Figure 1. Causal model (a) and belief distribution (b) for the two-coin example.**

The point to notice is that by specifying a causal model for predicting the outcome $E_2$, we automatically specified the variance of that prediction. In other words, when a person encodes probabilistic knowledge as a causal model of interacting variables, that person automatically specifies, not merely the marginal and joint distributions of the variables in the system, but also an entire set of future scenarios, describing how these probabilities would vary in response to future eventualities. It is this implicitly encoded dynamics that renders probabilistic statements random events, admitting distributions, intervals, and other confidence measures.

Why, then, is our confidence in $P(E_1) = 0.50$ higher than that of $P(E_2) = 0.50$? True, each of the mentioned contingencies might also hold in the case of a typical coin, pulled out of our pocket, however, because the probabilities associated with contingencies $C_2$ and $C_3$ are negligibly small, the resultant distribution of the assessment $P(E_1)$ will be extremely focussed about its mean 0.50. Thus, our confidence in the statement $P(E_1) = 0.50$ stems from not finding in our knowledge base a set of conditions which are both likely to happen and would substantially sway the assessment of $P(E_1)$ one way or the other.

There is, of course, a whole spectrum of conditions we can draw from trajectory calculations which would make the outcome of the coin sway in favor of either head or tail. For example, we can specify a narrow range of initial conditions (e.g., position, orientation, linear and rotational momentum) for which the outcome of tossing the coin is almost sure. However, such conditions would not qualify as meaningful contingencies because, unless one is concerned with scientific experimentation, the granularity and vocabulary required to specify these conditions lie beyond the level of abstraction common to everyday discourse. Abstraction levels are adopted to match the sort of evidence one expects to find under various circumstances and, since in normal coin-tossing settings one can envision no effective measurement nor hypothetical eventuality that would facilitate the fine physical distinctions necessary for determining the outcome of the toss, such distinctions were chosen to be glossed over and summarized in probabilistic terms. On the other hand, eventualities such as "the coin was loaded" do lie within the boundaries of normal discourse because they involve common actions and intentions which can be understood, envisioned and revealed by ordinary means e.g., the coin producer may confess to action or intention.

The granularity of the model chosen for predicting an event $E$ plays a major role in determining the balance between confidence and certainty. For example, had we chosen to explicate our entire knowledge about Newtonian mechanics and molecular physics within the knowledge base, there would be hardly room for probabilistic predictions. Rather, most predictions will be deterministic but, due to missing boundary conditions, they will be issued with zero level of confidence. In other words $P(E \mid c)$ would oscillate violently between 0 and 1 depending on the particular boundary condition assumed for $c$. Once we decide to exclude trajectory calculations from our knowledge base and summarize them by coarse-grain variables, probabilistic predictions become feasible and our confidence in these prediction is reflected by narrow distributions such as that shown in Figure 1(b).

Still, language granularity is not the only factor governing the level of confidence reflected in a model. The organizational structure connecting $E$ and $c$ plays no lesser role. Take, for example, the network of Figure 1(b). Its topology specifically proclaims $C_1$, $C_2$ and $C_3$ to be causal factors of $E$ and not the other way around, indicating that information should flow from the former to the latter and, hence, that $C_1$, $C_2$ and $C_3$ should serve as contingencies for $E$ and not vice versa. Had we chosen to express the identical joint probability distribution in a different graphical model, a different set of contingencies would be designated and different levels of confidence will ensue. For example, reversing the arrow between $E$ and $C$ and labling the arrow with the conditional probabilities $P(C_i | E)$ instead of the corresponding $P(E | C_i)$ would designate $E$ as a contingency for the $C_i$'s; the assessment $P(E) = 50\%$ will be issued with full confidence while the confidence of $P(C_i)$ will be measured by the variation between $P(C_i | E = true)$ and $P(C_i | E = false)$. Thus, aside from specifying the joint distribution of the variables in the systems and depicting their interdependencies, the choice of network structure carries semantics of its own. It reveals the procedures adopted by the model builder while encoding experiential data and those intended to be used in the retrieval of this data. In other words, the network structure designates which variables should constitute the *context* or *reference* for assessing the belief in other variables.

We are ready now to give this interpretation of confidence intervals a more formal underpinning, using $BEL(E)$ to denote $P(E \mid$ all evidence obtained so far$)$.

## A Formal Definition of Network-Induced Confidence Measures

Having agreed to associate partial confidence (in $BEL(E)$) with the susceptibility of $BEL(E | c)$ to the various contingencies in $C$ requires a definition of the contingency set $C$, relevant to $E$. Once $C$ is defined, we would be able to calculate the confidence in any probabilistic statement, say $BEL(E) = b$, directly from the network model in which $E$ is embedded. For example, we would be able to "simulate" the events in $C$ by instantiating various combinations of its variables, weighed by their appropriate probabilities, and measure the resulting fluctuations in $BEL(E | c)$.

Obviously, not every proposition would qualify for membership in the set $C$ of contingencies. De Finetti (1977) was careful to point out that, by cleverly manipulating $C$, one can fabricate any arbitrary distribution of $BEL(E | c)$, thus exhibiting any desirable confidence measure while maintaining the same uncertainty, $BEL(E)$. In the extreme case, if $C$ contains evidential variables which bear decisively on the truth of $E$, then the distribution of $BEL(E | c)$ will fluctuate violently between 0 and 1, indicating a total loss of confidence in the assessment of $BEL(E)$. Such behavior is not supported by introspective analysis. For example, suppose we were told that there is a bell hidden somewhere in the room, which will ring iff the coin turns up head; would that story alter our confidence in $P(E_1) = 0.50$? It should if the bell's sound $B$ is proclaimed to be a contingency relative to $E_1$. Yet, despite the fact that the conditional probability $P(E_1 | B)$ is extremely sensitive to whether $B$ is true or false, most people would agree that the story about the bell has no effect whatsoever on our confidence in the statement: $P(E_1) = 0.50$. Why? Apparently causal consequences of events do not qualify as contingencies for those events. Had the story been reversed, namely, that the outcome of the coin toss would, by some magical means, be influenced by an earlier sound of a bell, we would no doubt consider the bell part of the contingency set, affecting the confidence of $P(E_1)$.

Is it reasonable, then, to categorically exclude consequences from the contingency set? That seems to conflict with another mode of behavior where anticipated consequences play a dominant role in confidence judgment. Consider an example given in [Spiegelhalter, 1986]

> "--- a patient presented to a specialist may have a 10% chance of gastric cancer just from the known incidence in that referral clinic. However, one may be unwilling to make a decision until many further questions were asked, after which it may well be reasonable to perform an endoscopy even on the basis of the same 10% belief, since no further interrogation will substantially alter our belief."

Surely, the answers a patient gives to the specialist's questions are causal consequences of the patient conditions and the latter are causal consequences of the target event "gastric cancer". Yet, the mere option of obtaining answers to these questions renders the 10% belief less sure before the interrogation and more sure afterwards. Why? Spiegelhalter's justification "since no further interrogation will substantially alter our belief" may allude us to conclude that the set of possible answers are the very contingencies upon which our confidence rests. Would it be adequate, then, to identify the set $C$ with the set of observations one is likely to obtain?

Suppose a particular patient from the same referral clinic was stubbornly refusing to answer any questions, and his refusal was, in fact, so adamant that we are absolutely sure that "no further interrogation would alter our belief." Would this make us more confident in stating "$P(gastric\ cancer) = 10\%$" for this patient, compared with more cooperative ones (prior to their interrogation)? The answer is, of course, no. It is the diversity of beliefs about the patients conditions, not the availability of future observations that governs our confidence judgments.

The reason one is so sure in this case that interrogation will tighten the confidence in the assessment of $P$ (gastric cancer), is that interrogations often reveal alternative mechanisms that "explain away" previously observed symptoms and complaints, thus exonerating the target event $E$ = "gastric cancer" from direct responsibility. Since such alternative explanations play a dominant role in the assessment of $P$ (gastric cancer) they too (not the anticipated answers) should be included in the contingency set.

To summarize where we now stand on the issue of contingencies, the following features have been identified:

1.  Partial confidence in belief assessment for event $E$ is caused by anticipated fluctuations in a set $C$ of other events called contingencies and by the susceptibility of $BEL(E \mid c)$ to such fluctuations.

2.  Consequences of $E$ are excluded from $C$.

3.    Observing some consequences of $E$ often strengthens our confidence in the belief of $E$. Yet, whether or not we are about to gain access to such observations does not alter the confidence.

4.    After observing consequences of $E$ our confidence in estimating $BEL(E)$ depends on whether we can articulate alternative explanations to such observations, not involving $E$.

Matching these features to causal network terminology, the following coherent picture emerges.

1.    $C$ contains only direct parents of $E$ and direct parents of the observations at hand.

2.    Our confidence in the assessment of $BEL(E)$ is measured by the (narrowness of the) distribution of $BEL(E \mid c)$ as $c$ ranges over all combinations of contingencies, and each combination $c$ is weighed by its current belief $BEL(c)$.

3.    The effect of observations on the confidence attributed to $BEL(E)$ is three fold;

    3.1    Normally, the evidence observed renders $BEL(c)$ focused on a smaller number of combinations, thus minimizing the fluctuations of $BEL(E \mid c)$.

    3.2    Normally, the evidence observed renders $BEL(E \mid c)$ less sensitive to $c$ (e.g., $BEL(E \mid c)$ may rely more on the likelihood ratio $\lambda(E)$ and less on the prior $\pi(E)$)

    3.3    New observations may introduce new contingencies into $C$ (i.e., their direct parents), made up of alternative, uncertain explanations of the observations, which previously had no effect on $BEL(E)$.

The reason $C$ is composed of direct parents rather than more remote ancestors of the target events is that the former are identified as the most specific reference class [Kyburg, 1983] for determining the belief in the events under consideration. The reason we must qualify points 3.1 and 3.2 is that observational data may sometimes have the effect of weakening our confidence. For example, evidence attesting the fact that the coin producer has definitely attempted to tamper with its loading would undoubtedly weaken our confidence in the statement "$P(E_2) = 50\%$".

### An Example

To illustrate the approach, let us take an example from [Heckerman et al., 1987], originally given by [de Finetti, 1977]. Suppose we ask a person for his opinion on the probability that a given football team will win an upcoming game. We discover that he is uncomfortable supplying a point value because the probability for a win depends on three factors: (1) whether a star player, recently threatened with suspension by league authorities, will be able to play, (2) whether the playing field will be dry or muddy because of rain, and (3) whether the promise of a bonus to the winning players will be confirmed. We will label these events *Sus*, *Field* and *Bonus* respectively, and, assuming they are independent, we can represent their influence on the event *win* by the causal network of Figure 2. The contingency set $C$ consists of the direct parents of *win*, namely the variables *{Sus, Field, Bonus}*.

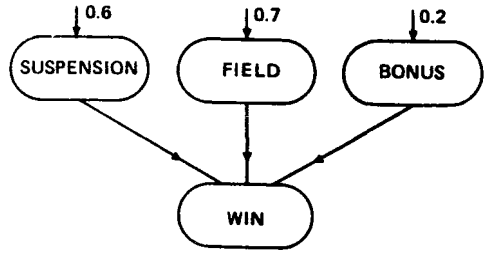

**Figure 2. Causal model for event** $E = Win$

Suppose the individual is able to assess point values for each of the influencing events, say $P$(Suspension) $= .6$, $P$(Dry-field) $= .7$, $P$(Bonus) $= .2$ as well as the probability of a win conditioned on each of the possible outcomes of the above influencing events. A hypothetical set of assessments are shown in the second column of Table 1. This completes the specification of the causal network and allows us to compute both the belief in the event *win* and the confidence in that belief.

The probability of each combination of outcomes is shown in the third column of Table 1. A plot of the third column versus the second column of Table 1 is shown in Figure 3. This plot graphically depicts the nature of partial confidence in the individual's assessment of $P$(Win). The quantity $P$(Win) varies as a function of the outcomes of the influencing events in the contingency set. In general, the wider the "curve" in such a plot, the lower the confidence in the assessment.

| Conditioning events | | | $P$(Win \| Events) | $P$(Events) |
|---|---|---|---|---|
| No-sus | Dry-field | Bonus | .7 | (.4)(.7)(.2) = .06 |
| No-sus | Dry-field | No-Bonus | .6 | (.4)(.7)(.8) = .22 |
| No-sus | Wet-field | Bonus | .6 | (.4)(.3)(.2) = .02 |
| No-sus | Wet-field | No-Bonus | .5 | (.4)(.3)(.8) = .10 |
| Sus | Dry-field | Bonus | .6 | (.6)(.7)(.2) = .08 |
| Sus | Dry-field | No-Bonus | .5 | (.6)(.7)(.8) = .34 |
| Sus | Wet-field | Bonus | .5 | (.6)(.3)(.2) = .04 |
| Sus | Wet-field | No-Bonus | .4 | (.6)(.3)(.8) = .14 |

**Table 1: The probability of a win conditioned on the influencing events.**

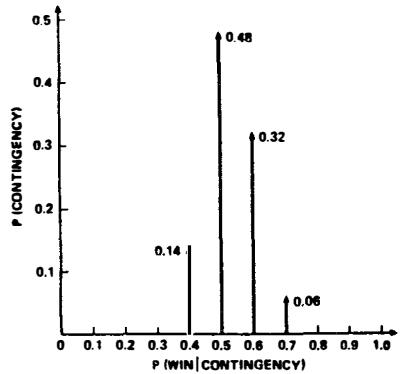

**Figure 3. Belief distribution associated with the model of Figure 2.**

The overall belief in a win is given by the mean of the distribution of Figure 3,

$$P(Win) = \sum P(Win \mid Sus, Field, Bonus) P(Sus, Field, Bonus) = .53.$$

where the sum ranges over all possible outcomes of the influencing events. The confidence interval can be measured by the range [.4, .7] or, better yet, by the standard deviation

$$\sigma = \left\{ \sum \left[ P(Win \mid Sus, Field, Bonus) - .53 \right]^2 P(Sus, Field, Bonus) \right\}^{\frac{1}{2}}$$

$$= \left[ (.4 - .53)^2 .14 + (.50 - .53)^2 .48 + (.6 - .53)^2 .32 + (.7 - .53)^2 .06 \right]^{\frac{1}{2}} = .0781$$

Now let us examine the effect of new evidence on the individual's belief and confidence. Assume that the individual is given the option to consult a member of the committee deciding the suspension of the star player. Additionally, he can call the weather bureau to get the weather forecast for the day of the match. These potential sources of information are represented by the two leaf nodes added to the network in Figure 4(a). Prior to consulting these sources, however, the distribution of $P(Win)$ remains the same as in Figure 3; the mere availability of information sources should not change the individual's confidence, since these do not belong to the contingency set $C$. On the other hand, assuming that the committee member assures us that the star player will not be suspended and that the weather report generates a likelihood ratio of 4:1 in favor of a wet-field, the belief distribution of the contingency set changes to $BEL(Suspension) = 0$, $BEL(Dry-field) = \dfrac{.7}{.7 + .3 \times 4} = .368$, $BEL(Bonus) = .2$ and the probabilities of a win are given by Table 2 and Figure 4(b).

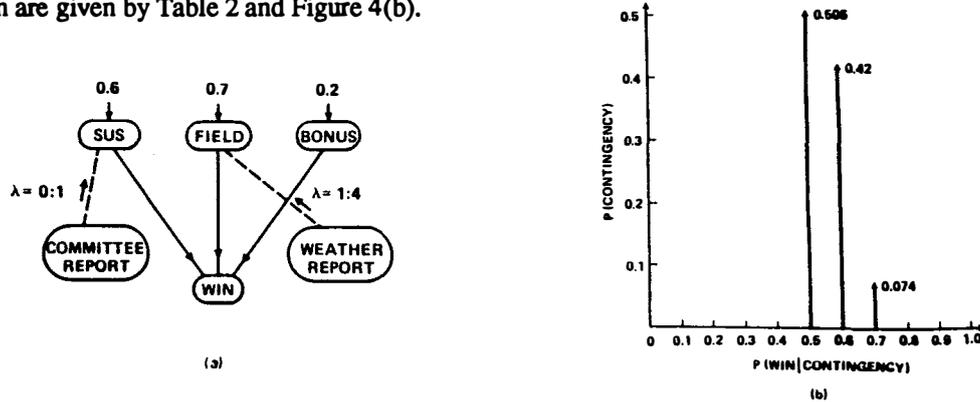

(a)

Figure 4. Causal model (a) and belief distribution (b) after obtaining reports.

| Conditioning Events | | | $BEL(Win \mid Events)$ | $BEL(Event)$ |
|---|---|---|---|---|
| No-Sus | Dry-field | Bonus | .7 | (1) (.368) (.2) = .074 |
| No-Sus | Dry-field | No-Bonus | .6 | (1) (.368) (.8) = .294 |
| No-Sus | Wet-field | Bonus | .6 | (1) (.632) (.2) = .126 |
| No-Sus | Wet-field | No-Bonus | .5 | (1) (.632) (.8) = .506 |

Table 2

Notably, the new distribution of $BEL(Win)$ is narrower, having the standard deviation of .0627 about the mean $BEL(Win) = .556$. This demonstrates the narrowing of confidence intervals by evidence that prunes and dampens the range of possible combinations in the contingency set $C$ (point 3.2 above).

To demonstrate the effect of an evidence that is a consequence of the central event *win* (point 3.1), ima-
gine that the individual involved could not attend the match and, having to leave town at that day, asked his brother
to call him if the home team wins. It is now two hours past the ending time of the game and his brother has not
called. This new piece of evidence, "no-call" is shown as a successor (i.e., consequence) of the variable *win* in the
causal network of Figure 5(a). Again, we allow the influencing events to range over the four possible value combi-
nations shown in Table 2, however, both the belief in each combination as well as the conditional belief in *win*,
given that combination, are now influenced by the evidence *no-call*.

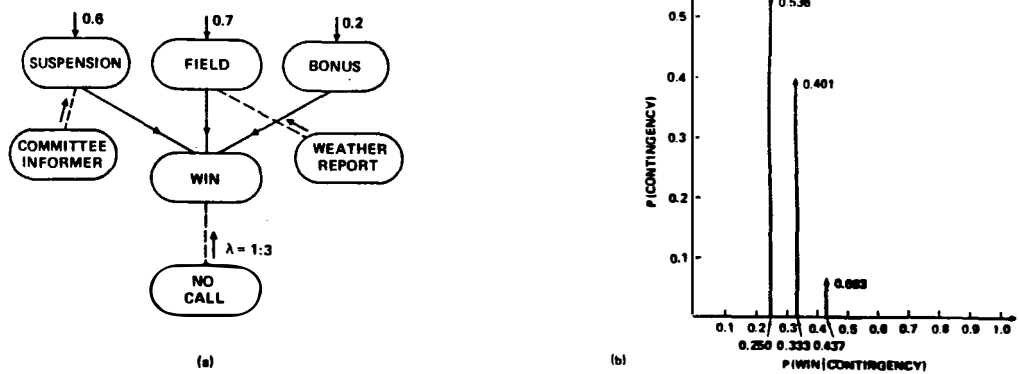

(a)                                              (b)

**Figure 5. Causal model (a) and belief distribution (b) after obtaining "no-call."**

Assuming that the new evidence *no-call* imparts a likelihood ratio of $\lambda = 3:1$ against *win*, the resulting be-
liefs are shown in Table 3.

| Event | | | $BEL(Win \mid Event)$ | $BEL(Event)$ |
|---|---|---|---|---|
| No-Sus | Dry-field | Bonus | .437 | .063 |
| No-Sus | Dry-field | No-Bonus | .333 | .281 |
| No-Sus | Wet-field | Bonus | .333 | .120 |
| No-Sus | Wet-field | No-Bonus | .250 | .536 |

Table 3

For example, $BEL(Win \mid Dry, Bonus)$ was calculated by

$$BEL(Win \mid Dry, Bonus) = \frac{.7}{.7 + .3 \times 3} = .437$$

and

$$BEL(Dry, Bonus) = \alpha \, \pi(Dry, Bonus) \, \lambda(Dry, Bonus) =$$

$$= \alpha \, .074(.7 \times 1 + .3 \times 3) = \alpha \, .1184 = (.1184 + .5292 + .2268 + 1.012)^{-1} \, .1184 = .063$$

We see that the new evidence tends to diminish the likelihood of *win* for every combination of the influencing
events and, simultaneously, it tends to "put-the-blame" on the combination (wet-field, no-bonus). The resulting
distribution of $BEL(win)$ is given in Figure 5(b), having mean .2953 and standard deviation .0542. This distribution
can be thought of as resulting from mental simulation of scenarios emanating from the contingency set, each
scenario being weighed and its impact assessed in light of the total evidence available.

To illustrate the introduction of new contingencies into $C$ (point 3.3) imagine that the individual involved
feels uncomfortable with the assessed likelihood ratio of $\lambda = 3:1$. Generally, such assessment would summarize a
large number of detailed possibilities explaining why the phone call did not materialize despite the event *win*. This
numerical summary would normally be adequate except that, in this particular situation, the individual is genuinely disturbed by the vivid possibility that he left behind the phone number of the wrong hotel. This new possibility should be described as a direct parent of the observation "no call" and will be treated as a new contingency. Under the assumption that the wrong phone number was given, the likelihood ratio would be 1:1 and the spread of $BEL(E \mid C_i)$ will be the same as in Figure 4(b). Under the assumption that the right phone number was given the likelihood ratio will, of course, be higher; perhaps 5:1 against $win$. The average over these two assumptions should coincide with Figure 5(b) but the overall spread of $BEL(E \mid e)$ can be much higher, depending on the probability attached to the event "wrong number." It is not uncommon to find one's sense of confidence vary at the whims of the assumptions one chooses to explicate at any given time.

In larger networks, the $BEL(e)$ can be computed using the chain-product rule of conditional probabilities, instantiating the variables in $C$ one at a time. A reasonable way to approximate the distribution of $BEL(E \mid e)$ would be to calculate only the 3-4 most likely combinations of contingencies from $C$ and examine how susceptible $BEL(E \mid e)$ is to these combinations. This would parallel people's tendency to justify lack of confidence by imagining a few likely scenarios leading to diverse consequences.

## Conclusions

In a recent paper, Kyburg (1987) makes a distinction between second order probabilities and probabilities of probabilities and concludes that the former adds nothing to the latter. In Kyburg's words: "so called second order probabilities have nothing to contribute conceptually to the analysis and representation of uncertainty," because "information about the accuracy of $P$ is fully expressed by a probability density function over $P$," [Cheeseman, 1985].

This paper offers an even stronger support to Kyburg's claim, showing that probabilities-of-probabilities, too, have nothing to contribute to the analysis and representation of uncertainty. Information about the accuracy of $P$ need not even be expressed by a "probability density function over $P$"; it is a built-in feature of the very same model that provides the information about $P$.

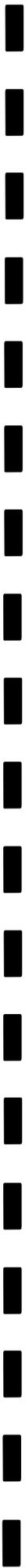